\documentclass[conference]{IEEEtran}
\IEEEoverridecommandlockouts
\usepackage{cite}
\usepackage{amsmath,amssymb,amsfonts}
\usepackage{algorithmic}
\usepackage{graphicx}
\usepackage{textcomp}
\usepackage{xcolor}
\def\BibTeX{{\rm B\kern-.05em{\sc i\kern-.025em b}\kern-.08em
    T\kern-.1667em\lower.7ex\hbox{E}\kern-.125emX}}
\begin{document}

\title{DisBeaNet: A Deep Neural Network to augment Unmanned Surface Vessels for maritime situational awareness \\
}

\author{\IEEEauthorblockN{1\textsuperscript{st} Srikanth Vemula}
\IEEEauthorblockA{\textit{Computer Science Department} \\
\textit{College of Saint Ben \& Saint Johns University}\\
Collegeville, USA \\
svemula001@csbsju.edu}
\and
\IEEEauthorblockN{2\textsuperscript{nd} Eulises Franco}
\IEEEauthorblockA{\textit{Computational Intelligence Lab (CIL))} \\
\textit{University of the Incarnate Word}\\
San Antonio, USA \\
francope@student.uiwtx.edu}
\and
\IEEEauthorblockN{3\textsuperscript{rd} Dr.Michael Frye}
\IEEEauthorblockA{\textit{Autonomous Vehice System Labs} \\
\textit{University of the Incarnate Word}\\
San Antonio, USA \\
mfrye@uiwtx.edu}
}

\maketitle

\begin{abstract}
Intelligent detection and tracking of the vessels on the sea play a significant role in conducting traffic avoidance in unmanned surface vessels(USV). Current traffic avoidance software relies mainly on Automated Identification System (AIS) and radar to track other vessels to avoid collisions and acts as a typical perception system to detect targets. However, in a contested environment, emitting radar energy also presents the vulnerability to detection by adversaries. Deactivating these Radiofrequency transmitting sources will increase the threat of detection and degrade the USV's ability to monitor shipping traffic in the vicinity. Therefore, an intelligent visual perception system based on an onboard camera with passive sensing capabilities that aims to assist USV in addressing this problem is presented in this paper. This paper will present a novel low-cost vision perception system for detecting and tracking vessels in the maritime environment. This novel low-cost vision perception system is introduced using the deep learning framework. A neural network, DisBeaNet, can detect vessels, track, and estimate the vessel's distance and bearing from the monocular camera. The outputs obtained from this neural network are used to determine the latitude and longitude of the identified vessel.
\end{abstract}


\section{Introduction}
In establishing reliable transportation in the sea, automation of vessel navigation and situational awareness place a vital role. Today, about 90\% of the world's cargo is seaborne[1]. Safeguarding the seaways is essential for goods and human lives, piracy, illegal fishery, ocean dumping, and refugee transportation. These reasons make maritime surveillance a significant element for government and private organizations to set up traffic monitoring for maritime environments. One such example that was set up by the European Maritime Safety Agency (EMSA) is to monitor vessel traffic and receive information on ships and their movements and notify if there are any dangerous cargos [2]. One major issue with maritime surveillance is the immense area of the sea on the earth's surface which makes monitoring of vessel traffic a complex task [3]. The only approach to obtain reliable information about a vessel's current position is the Automatic Identification System (AIS) [4]. There are also other approaches like radar perception systems and fusion perception systems that can track the vessel to avoid collisions.

One such approach that can improve maritime domain awareness is the Earth Observation(EO) satellite data which proves to be a valuable source of information. Numerous significant efforts are made in research using optical and radar satellite images [5-7]. However, in most cases, the gathered images are analyzed long after the data is acquired [8]. To address this bottleneck, near real-time services came into the existence, and progress has been made, which are by far the best resource that provides information is measured at a range of 15 min from on-ground reception [9,10]. However, there is still a significant amount of time delay between the data acquisition onboard and the amount of data received from on-ground.  On the other hand, the image data size is large, and their down-link requires direct contact to a ground station. The delay in acquiring the data can be from hours to even days[11].

Another constraint in using these EO satellites is their inability to monitor a defined region of interest for critical applications continuously. Typically, for a vessel detection carried out using satellite data, there is a reasonable spatial resolution in Low Earth Orbit (LEO) with a speed of approximately 7 km/s over the ground and a revisit cycle of several days [12].

Unmanned autonomous vehicles [13] have emerged as a promising option in providing maritime surveillance services. One such instance regarding this is the  Remotely Piloted Aircraft Systems (RPAS), which is operated by the European Maritime Safety Agency [14] that operates several services required by maritime surveillance. Due to its small, lightweight, and ability to take off within minutes [13]. However, there are limitations regarding their operational flight duration and the range of their limited geographical applicability. High-altitude pseudo-satellites (HAPS) is a perfect fit for long-range endurance monitoring tasks, but it is still in the initial stages of development. One of the most famous HAPS is the Airbus Zephyrs, which can carry a payload of up to 20 kg and can hold nearly 26 days which holds the world record for longest uninterrupted flight [15]. Even though there is a huge possibility and flexibility in deploying these HAPS in remote areas, the satellites still face a bottleneck.

Another important tasks in autonomous system during vessel navigation is aimed at increasing safety and efficiency. In obtaining this one crucial task that involves in autonomous obstacle detection recently is looking for the combination of environment perception sensors. In contrast, vision-based obstacle detection is still considered irreplaceable[16]. In recent times, computer vision technology is increasingly getting more attention, and most of the data obtained by it is used compared to the radar and LiDAR [17]. Whereas in computer vision, obstacle detection is mostly carried out using stereo vision. Images from two stereo cameras are used to triangulate and estimate the distances of the objects detected and determine potential obstacles viewed by the cameras [18]. Stereo vision is used as a single system for detecting obstacles and is used in combination with other sensors. One such instance would be an obstacle detection system developed in [19] that uses a combination of computer vision and laser scanners. In which the laser provides the point cloud from which the system extracts the obstacles. These extracted clusters are used for generating both regions of interest and obstacle classification using machine learning.

Apart from these stereo vision techniques, there are also monocular cues used for texture, gradient, defocus, and color/haze, which provide crucial depth information while conducting obstacle detection. Some of these cues are even applied to regions where the stereo vision is not good. Due to this, some researchers pursue the idea of human perception of depth by combining stereo and monocular cues. It was shown that by combining this more accurate depth estimation can be determined. However, distance and bearing estimation from a single camera is difficult just by using the passive sensing capabilities of the camera. The problem of distance and bearing estimation from a single camera of a variety of structured or unstructured environments in both indoor and outdoor has not been attempted to the best of the researchers' knowledge. Thus, in this paper, a novel method for vessel distance and bearing estimation from a single camera that does not require any prior knowledge about the scene or explicit knowledge about various camera parameters is presented. The presented vessel distance and bearing estimation vision perception system based on a multi-hidden layer neural network, called DisBeaNet, is presented. The estimated distance and bearing from this DisBeaNet are used to predict in generating Geo-referenced tracks of the detected vessel.

In this paper, initial results of detected vessel distance estimation and bearing from a monocular camera are demonstrated using a novel deep learning architecture called DisBeaNet – a neural network that is responsible for estimating the distance and bearing of a detected vessel in the sea. From which the Geo-referenced tracks of the detected vessel is calculated. This proposed low-cost intelligent visual perception system is developed  mainly targeting autonomous vessels in maritime environments. Furthermore, the paper is structured as follows; first, the author presents the II. Overview of the system, then moves on to talk about the III. Experimental Results from the developed low-cost intelligent vision perception system and follows with the IV.conclusion and future work.

\section{Overview of the System}
The overview of the system architecture of novel low-cost intelligent visual perception system for detecting vessel and predicting the Geo-referencing tracks of the vessel in an open sea is illustrated in Fig. 1. In this input images are taken from the monocular camera and is fed to the modified state-of-the-art computer vision object detection algorithm YOLO (You Only Look Once) [20] which is trained on custom vessel dataset. Obtaining accuracy and speed is important in visually tracking system for such reasons a modified YOLOv3 with tracking functionality is used in this system. Once the model is trained on the custom vessel dataset the optput of the model is going to be bounding boxes of the detected object and the class name. The resulted bounding boxes parameters along with the distance and bearing acts as input features to be trained for the DisBeaNet neural network and outputs the distance and bearing of the detected vessel. After obtaining these parameters from DisBeaNet they are fed to the system where the Geo-referenced tracks of the vessel detected are calculated with the corresponding latitude and longitude of the vessel.An illustration of the workflow process of the system is presented below in the Fig.1.

\begin{figure}
    \centering
    \includegraphics[width=0.5\textwidth]{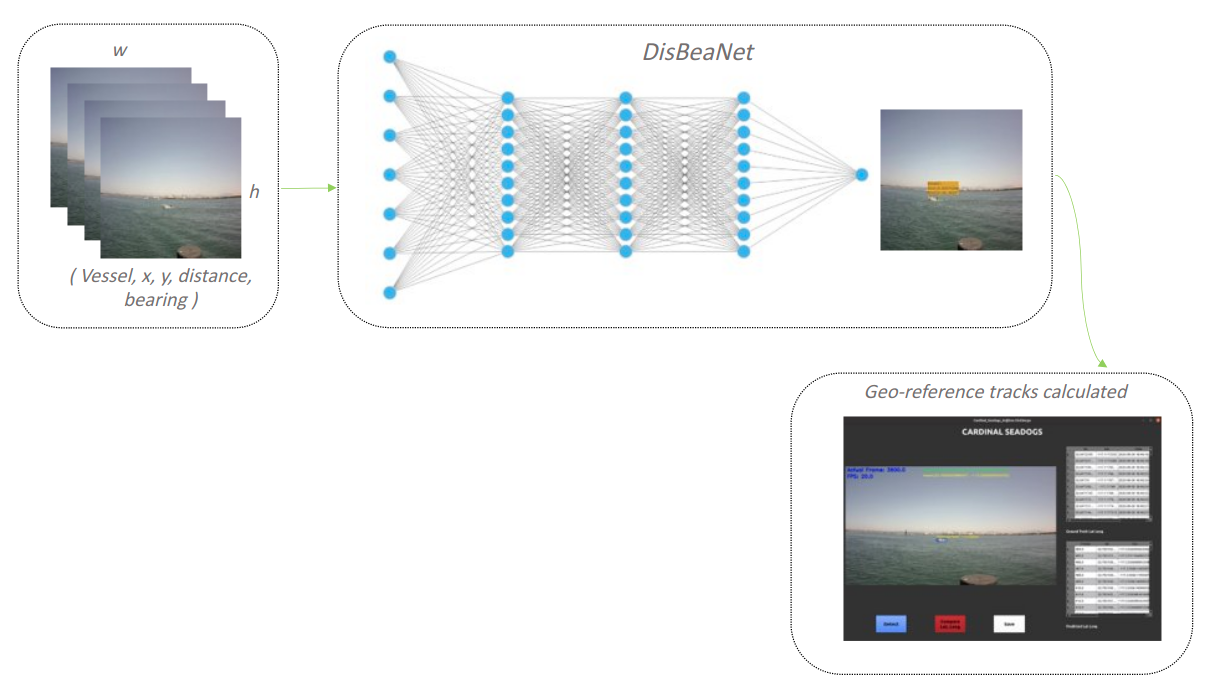}
    \caption{The DisBeaNet -based system used for  distance and bearing estimation of the detected vessel for predicting the Geo-referencing tracks of the vessel from a monocular camera}
    \label{fig:label}
\end{figure}

In the following sections each stage that makes up the entire system architecture is described in detailed. Initially, section demonstrates the process involved in preparing the custom vessel dataset for this task. In which data is collected from AI tracks at sea competition which consists of videos of the vessel traveling in the open sea and the ground truth values of the vessel distance, bearing and their Geo-referenced tracks are obtained. From the videos each individual frames are separated and are used for annotating the data using labelme[21] annotating tool. After annotating the data from the videos and constructing the dataset in COCO format and then move on to start training the DisBeaNet.In training the DisBeaNet, a supervised learning technique was used and the details of the structure and training of DisBeaNet neural network is presented in the following section of this paper.

\subsection{Real-time Vessel Detection and Tracking}

In this system, the first step is to construct the custom vessel dataset, which will serve as the vital data component for the rest of the system to work. The videos were acquired from the AI tracks at Sea challenge conducted by DoD, from which ground truth values of the distance, bearing, latitude, and longitude were obtained. Once the data was acquired from the competition vessel, data is annotated from the videos using the labelme [21] data annotation tool. After constructing the custom vessel dataset, a modified YOLO object detection state-of-the-art algorithm was used to detect and track the vessel in an open sea environment using a monocular camera. These predictions are encoded as S x S x (B x 7 + C) tensor, which denotes that the image that is provided to the modified algorithm is divided into S x S splits. Each splits consists of a B bounding boxes and is represented by its 7 location parameters: x, y, w, h,vessel,distance,and bearing.Another feature C is also predicted that indicates the class label of each bounding box predicted in the frame. The visual features captured from the modified YOLO algorithm confidence feature are nullified for the visual tracking that end up having seven features act as inputs for the DisBeaNet neural network, which will be explained in detail in the next section of this paper.

\subsection{DisBeaNet:  A Deep Neural Network to augment Unmanned Surface Vessels in predicting Distance and Bearing using Monocular Camera}

In this section of the paper demonstrates on how the DisBeaNet neural network was created and trained using a custom vessel dataset. The main objective of this DisBeaNet  neural network is to estimate the distance and bearing of the detected vessel from the passive sensing capabilities of the monocular camera. The structure of this DisBeaNet neural network comprised of three layers; input layer, hidden layer, and output layer. The input layer consists of seven neurons that acts as inputs for the neural network. When coming to the number of hidden layers (1,2,3,5 and 20) experiments were conducted to measure the accuracy of the distance and bearing estimation in over 120k epochs based on a trial and error. Out of each a closest distance and bearing estimation  accuracy is achieved when three hidden layers was used and there is not huge difference observed when compared to 2,3,5, and 20 hidden layers. Due to these reasons three hidden layers were chosen for this DisBeaNet neural network.Finally one output layer is used to capture the features that includes  x, y, w, h, distance,bearing and class label. The structure of the DisBeaNet neural network is presented in the Fig. 2.

\begin{figure}
    \centering
    \includegraphics[width=0.5\textwidth]{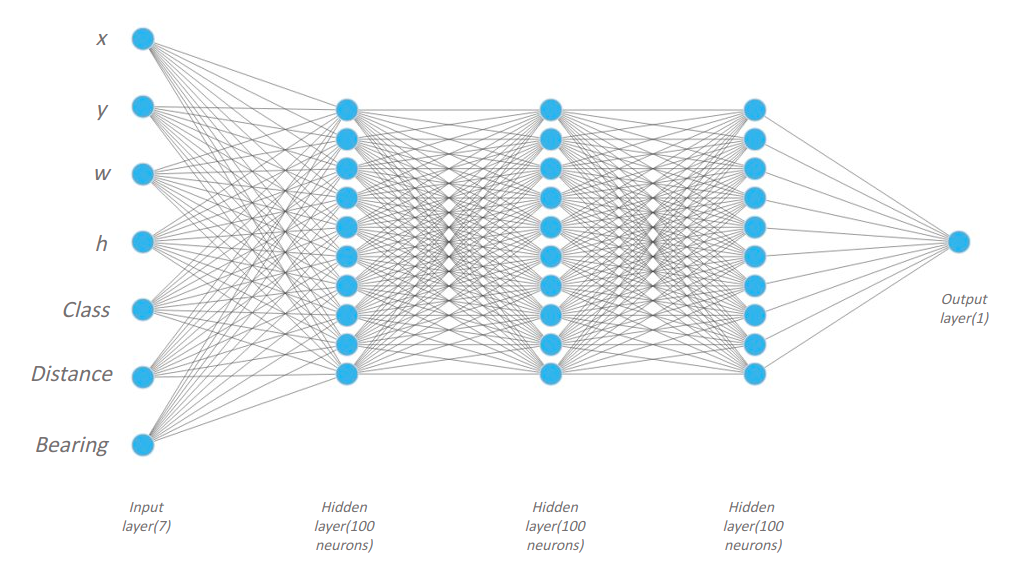}
    \caption{DisBeaNet Neural Network for estimating Distance and bearing of the Vessel Detected}
    \label{fig:label}
\end{figure}

For training this neural network a custom vessel dataset is prepared from 3000 images curated from the videos and the ground truth data that is obtained in the previous stage from the competition. After the preparation of the dataset training was carried out for seven days to obtain 94\% accuracy on Nvidia GeFroce RTX 3060ti using TensorFlow and Keras. After finishing the training, the model is then used in the Geo-referenced tracks predicting system by taking these parameters from the DisBeaNet.The results from the DisBeaNet that were obtained are presented in the below Fig. 3.

\begin{figure}
    \centering
    \includegraphics[width=0.5\textwidth]{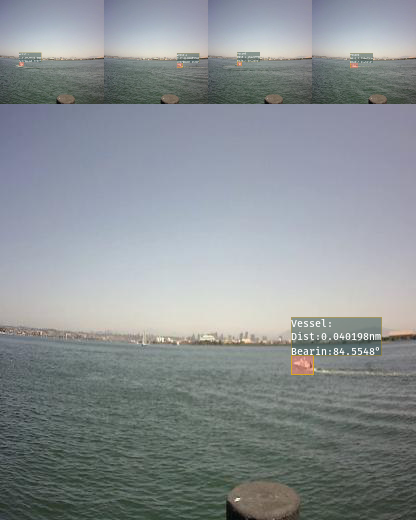}
    \caption{DisBeaNet Neural Network results for estimating distance \& Bearing of the detected vessel}
    \label{fig:label}
\end{figure}

The next section of the paper demonstrates the vessel Geo-referenced tracking system where the predicted latitude and longitude of the detected vessel from DisBeaNet model.

\subsection{Geo-Referenced Vessel Predicting System}

In this section, the final stage of the system, is discussed where the distance and bearing parameters are used to calculate the latitude and longitude of the detected vessel.The system used to predict the latitude and longitude is presented in Fig. 4.
\begin{figure}
    \centering
    \includegraphics[width=0.5\textwidth]{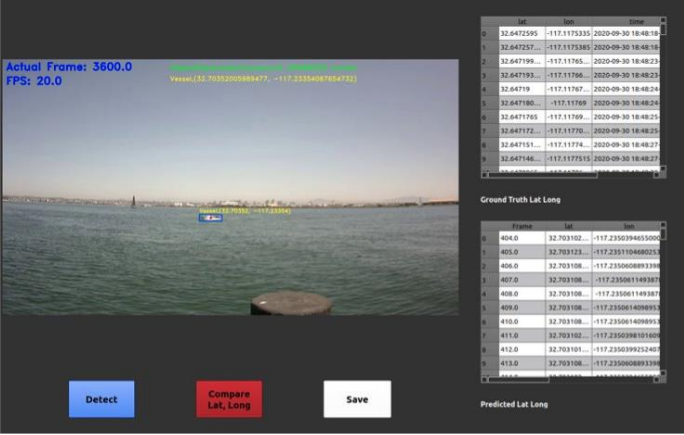}
    \caption{Vessel Geo-Referenced Predicting System}
    \label{fig:label}
\end{figure}

In this system the calculation of the predicted latitude and longitude of the vessel carried out by taking into consideration the parameters like radius of the earth R, predicted distance and bearing obtained from the DisBeaNet  neural network  and stored under the variable names brng, and dNM. By taking these parameters as input along with the camera latitude and longitude. The latitude and longitude are calculated for the vessel detected in real-time. The pseudo code that performs these operations are presented in the Fig. 5. After obtaining these predicted values from the DisBeaNet neural network the Geo-Referenced tracks are predicted.

\begin{figure}
    \centering
    \includegraphics[width=0.5\textwidth]{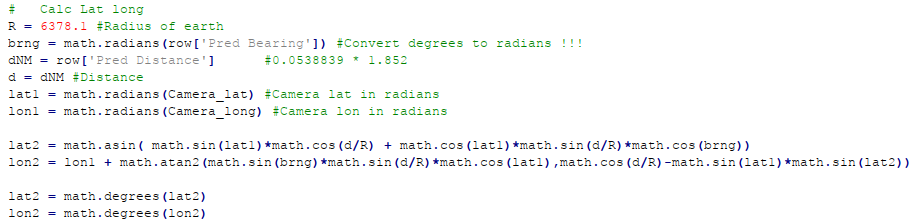}
    \caption{Pseudo code for predicted latitude and longitude}
    \label{fig:label}
\end{figure}

\section{Results}
In this section of the paper the experimental results of the DisBeaNet neural network results are evaluated with the ground truth values using root mean square error (rmse) and is calculated using the formula stated below.
 				
$$
rmse = \sqrt{\Sigma_{i=1}^{n}{\Big({Predicted_i -Actual_i}\Big)^2}\over{N}}
$$

As from the plots it shows that the rmse is very small in terms of the distance and bearing from DisBeaNet neural network which is quite promising.The visual representation of the predicted and ground truth values of the distance of the vessel from the camera along with the root mean square error can be seen Fig. 6.

\begin{figure}
    \centering
    \includegraphics[width=0.4\textwidth]{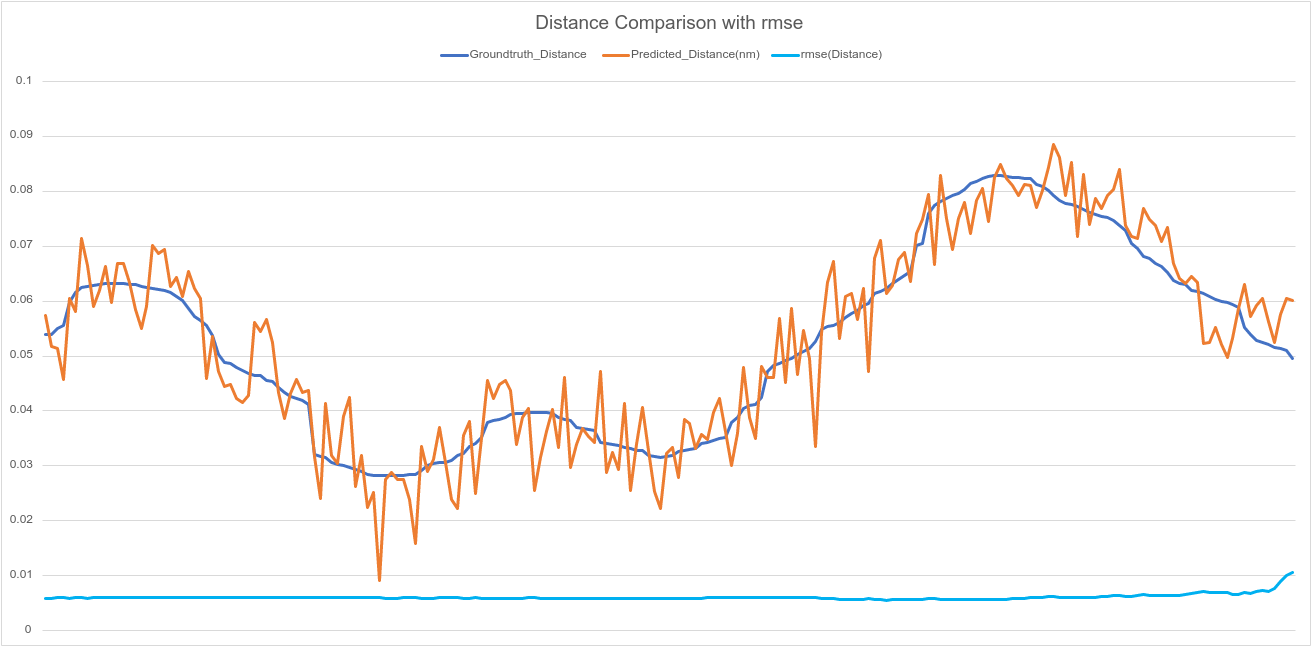}
    \caption{Distance Comparison between Predicted and Ground-truth values from DisBeaNet}
    \label{fig:label}
\end{figure}

From the figure it can be noticed that the rmse is negligible. When plotted the estimated bearing from the neural network the visual representation of the predicted and ground truth values of the bearing from the DisBeaNet is presented in the Fig. 7 along with its rmse values.

\begin{figure}
    \centering
    \includegraphics[width=0.4\textwidth]{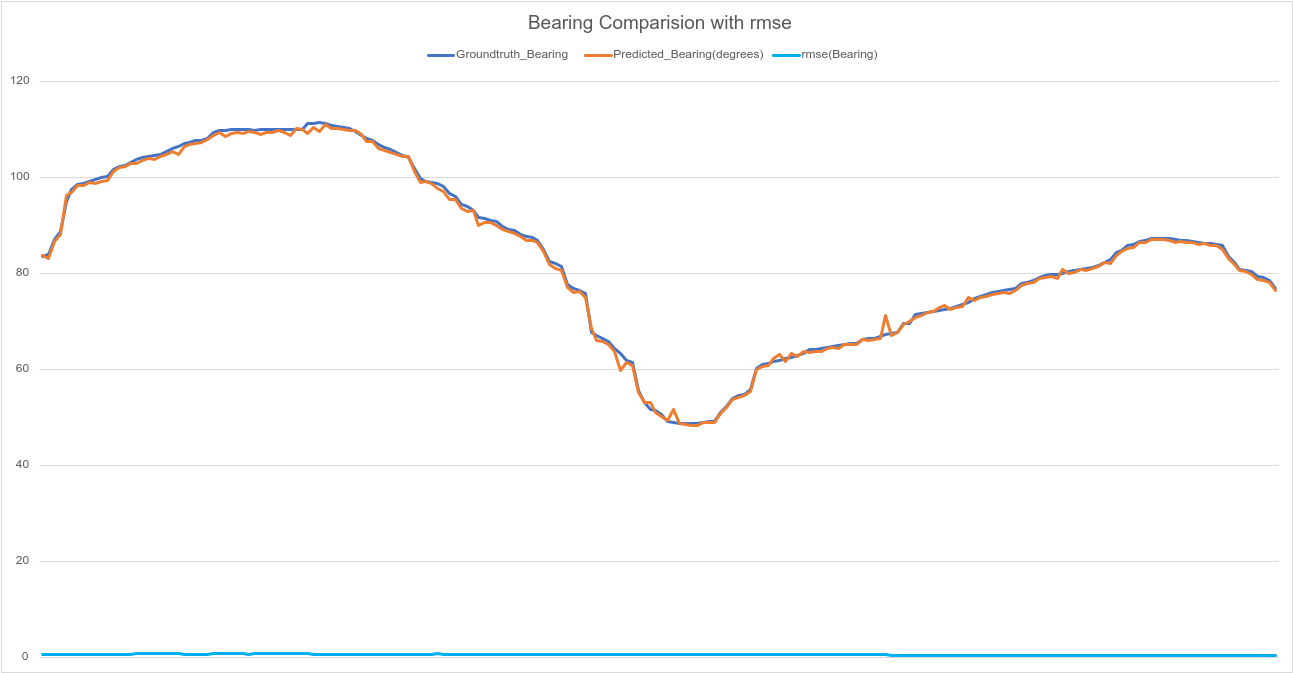}
    \caption{Bearing Comparison between Predicted and Ground-truth values from DisBeaNet}
    \label{fig:label}
\end{figure}

It is also noticed that when compared between the ground truth and predicted values of both the latitude  and longitude calculated from the predicting system that is seen in Fig. 8 and 9.

\begin{figure}
    \centering
    \includegraphics[width=0.4\textwidth]{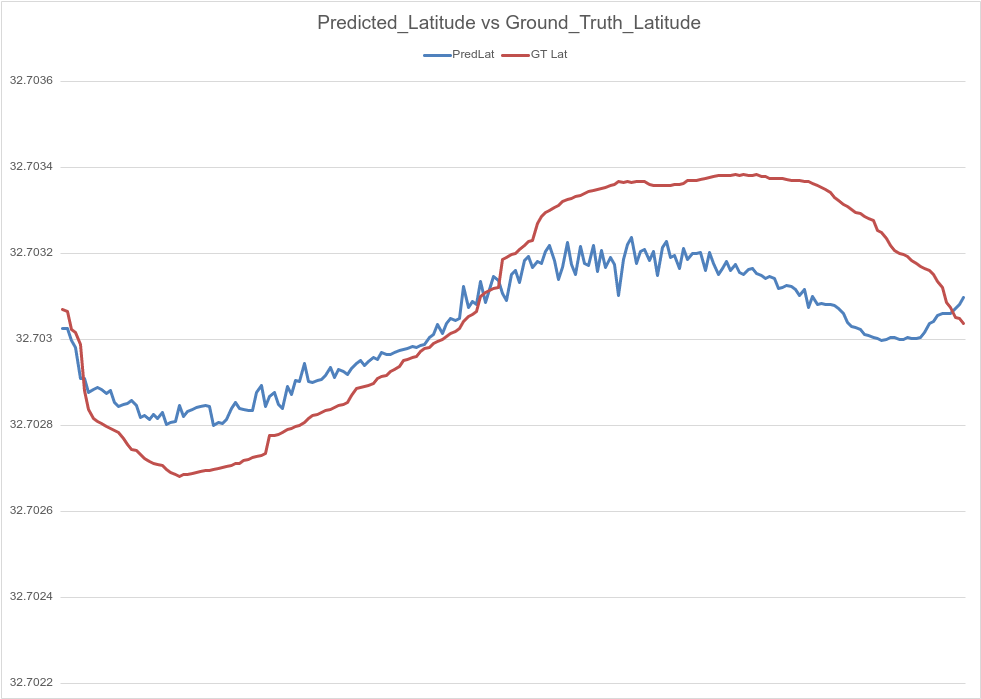}
    \caption{Comparison of Predicted Lat, Long with the Ground Truth Lat}
    \label{fig:label}
\end{figure}

\begin{figure}
    \centering
    \includegraphics[width=0.4\textwidth]{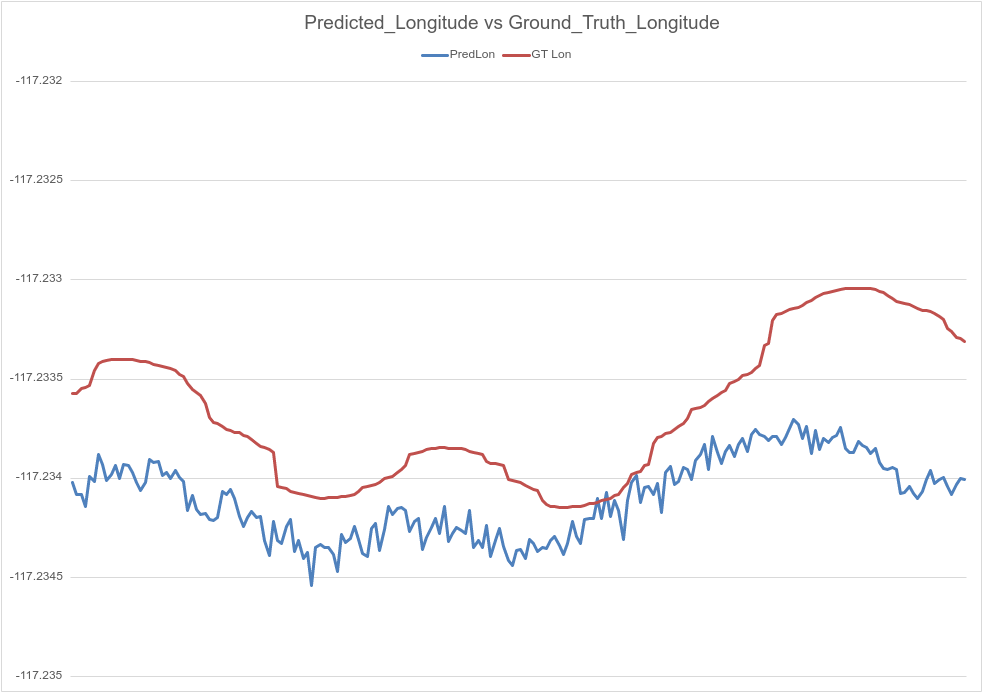}
    \caption{Comparison of Predicted Lat, Long with the Ground Truth Long}
    \label{fig:label}
\end{figure}

Shows that in the predicted latitude and longitude values the difference in minutes and seconds values has been noticed by the researchers and will take into consideration for the future work which will be discussed in the next section.

\section{Conclusion and Future Work}\label{AA}
After noticing encouraging results with the DisBeaNet  and at the same time noticeable difference between the ground truth and predicted latitude and longitude in terms of minutes and seconds from the predicting system. The researchers plan to look into this in the future work and make the system more reliable and reduce the amount of dependencies in conducting traffic avoidance in unmanned surface vehicles(USV) in maritime environment by using Deep Learning and computer vision. Apart from that, the current novel low-cost vision perception system that was developed shows promising results that will encourage to look into this vessel detection and tracking capabilities for maritime technology in USV's.

\section*{Acknowledgment}

This research was partially supported by grant Proposal W911NF2010260 from the Army Research Office.

\vspace{12pt}

\end{document}